\documentclass[11pt,letterpaper]{article}
\usepackage{hyperref}
\usepackage{emnlp2016}
\usepackage{times}
\usepackage{latexsym}
\usepackage{tikz}
\usepackage{enumitem}
\usepackage[font=footnotesize]{caption}
\usepackage{subcaption}
\usepackage{booktabs}
\usepackage{multicol}
\usepackage{amsmath}
\usepackage{amssymb}
\usepackage{amsthm}
\usepackage[framemethod=TikZ]{mdframed}
\usepackage{xcolor}
\usepackage[export]{adjustbox}

\newcommand{\szero}{\textrm{S}0}
\newcommand{\lzero}{\textrm{L}0}
\newcommand{\sone}{\textrm{S}1}

\newcommand\numberthis{\addtocounter{equation}{1}\tag{\theequation}}

\setlist{noitemsep}

\definecolor{highlight}{RGB}{197,59,35}

\emnlpfinalcopy
\def\emnlppaperid{***}

\newmdenv[%
  roundcorner=2pt,
  topline=false,
  bottomline=false,
  leftline=false,
  rightline=false,
  backgroundcolor=gray!16
]{samplebox}

\title{Reasoning about Pragmatics with Neural Listeners and Speakers}

\author{Jacob Andreas \and Dan Klein \\
      Computer Science Division \\
      University of California, Berkeley \\
      {\tt \{jda,klein\}@cs.berkeley.edu}}

\date{}

\begin{document}

\maketitle

\begin{abstract}
  We present a model for contrastively describing scenes, in which
  context-specific behavior results from a combination of inference-driven
  pragmatics and learned semantics.  Like previous learned approaches to
  language generation, our model uses a simple feature-driven architecture (here
  a pair of neural ``listener'' and ``speaker'' models) to ground language in
  the world. Like inference-driven approaches to pragmatics, our model actively
  reasons about listener behavior when selecting utterances. For training, our
  approach requires only ordinary captions, annotated \emph{without}
  demonstration of the pragmatic behavior the model ultimately exhibits. In
  human evaluations on a referring expression game, our approach succeeds 81\%
  of the time, compared to 69\% using existing techniques. 
\end{abstract}

\section{Introduction}

We present a model for describing scenes and objects by reasoning about context
and listener behavior. By incorporating standard neural modules for image
retrieval and language modeling into a probabilistic framework for pragmatics,
our model generates rich, contextually appropriate descriptions of structured
world representations.

\begin{figure}
  \centering
  \hfill
  \includegraphics[cfbox=highlight 4mm 0mm,width=0.4\columnwidth]{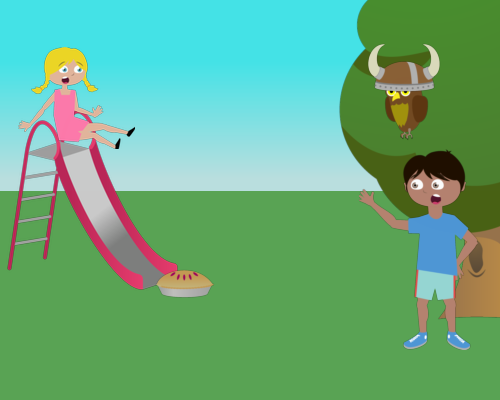}
  \hfill
  \includegraphics[cfbox=white 4mm 0mm,width=0.4\columnwidth]{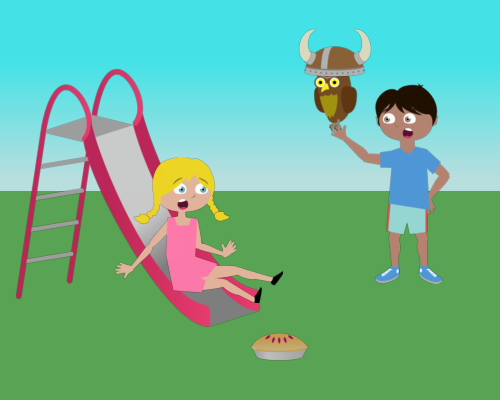}
  \hfill
  \strut
  \\
  {\footnotesize \strut \hspace{1.2cm} (a) target \hspace{2.4cm} (b) distractor
  \hfill \strut }\\[1em]
  \begin{samplebox}
    \centering
    \small
    \emph{the owl is sitting in the tree} 
  \end{samplebox}
  {\footnotesize (c) description}\\[0.5em]
  \caption{Sample output from our model. When presented with a target
    image (a) in contrast with a distractor image (b), the model generates a
    description (c). This description mentions a \emph{tree}, the distinguishing
    object present in (a) but not in (b), and situates it with respect to other
    objects and events in the scene.}
    \vspace{-1em}
  \label{fig:teaser}
\end{figure}

This paper focuses on a \emph{reference game} RG played
between a listener $L$ and a speaker $S$. \\[-0.8em]
\begin{equation}
  \fontsize{9.5pt}{11pt}\selectfont
  \tag{RG}
  \strut
  \hspace{-3em}
  \begin{aligned}
      \parbox{6.5cm}{
        \begin{enumerate}
          \setlength\itemsep{0.6em}
          \item Reference candidates $r_1$ and $r_2$ are revealed to both players.
          \item $S$ is secretly assigned a random target
            $t \in \{1, 2\}$.
          \item $S$ produces a description $d = S(t, r_1, r_2)$, which is
            shown to $L$.
          \item $L$ chooses $c = L(d, r_1, r_2)$.
          \item Both players win if $c=t$.
        \end{enumerate}
      }
  \end{aligned}
\end{equation}
\autoref{fig:teaser} shows an example
drawn from a standard captioning dataset \cite{Zitnick14Abstract}.

In order for the players to win, $S$'s description $d$ must be \emph{pragmatic}:
it must be informative, fluent, concise, and must ultimately encode an
understanding of $L$'s behavior.  In \autoref{fig:teaser}, for example,
\emph{the owl is wearing a hat} and \emph{the owl is sitting in the tree} are
both accurate descriptions of the target image, but only the second allows a
human listener to succeed with high probability. RG is the focus of many papers
in the computational pragmatics literature: it provides a concrete
generation task while eliciting a broad range of pragmatic behaviors, including
conversational implicature \cite{Benotti09Implicature} and context dependence
\cite{Smith13BayesianPragmatics}.  Existing computational models of pragmatics
can be divided into two broad lines of work, which we term the \emph{direct} and
\emph{derived} approaches.  

Direct models (see \autoref{sec:related} for examples) are based on a
representation of $S$. They learn pragmatic behavior by example. Beginning with
datasets annotated for the specific task they are trying to solve (e.g.\
examples of humans playing RG), direct models use feature-based architectures to
predict appropriate behavior without a listener representation. While quite
general in principle, such models require training data annotated specifically
with pragmatics in mind; such data is scarce in practice.

Derived models, by contrast, are based on a representation of $L$.  They first
instantiate a \emph{base listener} $\lzero$ (intended to simulate a na\"ive,
non-pragmatic listener). They then form a \emph{reasoning speaker} $\sone$,
which chooses a description that causes $\lzero$ to behave correctly.  Existing
derived models couple hand-written grammars and hand-engineered listener models
with sophisticated inference procedures. They exhibit complex behavior, but are
restricted to small domains where grammar engineering is practical.

The approach we present in this paper aims to capture the best aspects of both
lines of work. Like direct approaches, we use machine learning to acquire
a complete grounded generation model from data, without domain knowledge in the
form of a hand-written grammar or hand-engineered listener model. But like
derived approaches, we use this learning to construct a \emph{base} model, and
embed it within a higher-order model that reasons about listener responses.  As
will be seen, this reasoning step allows the model to make use of weaker
supervision than previous data-driven approaches, while exhibiting robust
behavior in a variety of contexts. 

Our goal is to build a derived model that scales to real-world datasets without
domain engineering.  Independent of the application to RG, our model also
belongs to the family of neural image captioning models that have been a popular
subject of recent study \cite{Xu15SAT}.  Nevertheless, our approach appears to
be:
\begin{itemize}
  \item the first such captioning model to reason explicitly about listeners
    \\[-0.7em]
  \item the first learned approach to pragmatics that requires only
    \emph{non-pragmatic} training data
\end{itemize}

Following previous work, we evaluate our model on RG, though the general
architecture could be applied to other tasks where pragmatics plays a core role.
Using a large dataset of abstract scenes like the one shown in
\autoref{fig:teaser}, we run a series of games with humans in the role of $L$
and our system in the role of $S$. We find that the descriptions generated by
our model result in correct interpretation 17\% more often than a recent learned
baseline system.  We use these experiments
to explore various other aspects of computational pragmatics, including
tradeoffs between adequacy and fluency, and between computational efficiency and
expressive power.\footnote{Models, human annotations, and code to generate all
tables and figures in this paper can be found at 
\url{http://github.com/jacobandreas/pragma}.}

\section{Related Work}
\label{sec:related}

\paragraph{Direct pragmatics}

As an example of the direct approach mentioned in the introduction,
\newcite{FitzGerald13Referring} collect a set of human-generated referring
expressions about abstract representations of sets of colored blocks. Given a
set of blocks to describe, their model directly learns a maximum-entropy
distribution over the set of logical expressions whose denotation is the target
set. Other research, focused on referring expression generation from a
computer vision perspective, includes that of \newcite{Mao15Generation} and
\newcite{Kazemzadeh14ReferIt}.

\paragraph{Derived pragmatics}

Derived approaches, sometimes referred to as ``rational speech acts'' models,
include those of \newcite{Smith13BayesianPragmatics}, \newcite{Vogel13Grice},
\newcite{Golland10Game}, and \newcite{Monroe15RationalSpeech}.  These couple
template-driven language generation with probabilistic or game-theoretic
reasoning frameworks to produce contextually appropriate language: 
intelligent listeners reason about the behavior of reflexive speakers, and even
higher-order speakers reason about these listeners.  Experiments
\cite{Frank09PragmaticExperiments} show that derived approaches explain human
behavior well, but both computational and representational issues restrict their
application to simple reference games. They require domain-specific
engineering, controlled world representations, and pragmatically annotated
training data.

An extensive literature on computational pragmatics considers its application to
tasks other than RG, including instruction following \cite{Anderson91MapTask}
and discourse analysis \cite{Jurafsky97PragmaticDiscourse}.

\paragraph{Representing language and the world}

In addition to the pragmatics literature, the approach proposed in this paper
relies extensively on recently developed tools for multimodal processing of
language and unstructured representations like images. These includes both image
retrieval models, which select an image from a collection given a textual
description \cite{Socher14Multimodal}, and neural conditional language models,
which take a content representation and emit a string \cite{Donahue15LRCN}.

%

\section{Approach}

Our goal is to produce a model that can play the role of the speaker $S$ in RG.
Specifically, given a target referent (e.g.\ scene or object) $r$ and a
distractor $r'$, the model must produce a description $d$ that uniquely
identifies $r$. For training, we have access to a set of
\emph{non-contrastively} captioned referents $\{(r_i, d_i)\}$: each training
description $d_i$ is generated for its associated referent $r_i$ in isolation.
There is no guarantee that $d_i$ would actually serve as a good referring
expression for $r_i$ in any particular context. We must thus use the training
data to ground language in referent representations, but rely on reasoning to
produce pragmatics.

Our model architecture is compositional and hierarchical. We begin
in \autoref{sec:approach:modules} by describing a collection of ``modules'':
basic computational primitives for mapping between referents, descriptions, and
reference judgments, here implemented as linear operators or small neural
networks.  While these modules appear as substructures in neural architectures
for a variety of tasks, we put them to novel use in constructing a reasoning
pragmatic speaker.

\autoref{sec:approach:base} describes how to assemble two base models: a
\emph{literal speaker}, which maps from referents to strings, and a
\emph{literal listener}, which maps from strings to reference judgments.
\autoref{sec:approach:reasoning} describes how these base models are used to
implement a top-level \emph{reasoning speaker}: a learned, probabilistic,
derived model of pragmatics.  

\subsection{Preliminaries}
\label{sec:approach:prelim}

Formally, we take a description $d$ to consist of a sequence of words $d_1,
d_2, \ldots, d_n$, drawn from a vocabulary of known size. For encoding, we
also assume access to a feature representation $f(d)$ of the sentence (for
purposes of this paper, a vector of indicator features on $n$-grams). These two
views---as a sequence of words $d_i$ and a feature vector $f(d)$---form the
basis of module interactions with language.

Referent representations are similarly simple.  Because the model never
generates referents---only conditions on them and scores them---a vector-valued
feature representation of referents suffices.  Our approach is completely
indifferent to the nature of this representation. While the experiments in this
paper use a vector of indicator features on objects and actions present in
abstract scenes (\autoref{fig:teaser}), it would be easy to
instead use pre-trained convolutional representations for referring to natural
images. As with descriptions, we denote this feature representation
$f(r)$ for referents.

\subsection{Modules}
\label{sec:approach:modules}

\begin{figure}
  \centering
  \includegraphics[width=\columnwidth, trim=1cm 5cm 7.5cm 1.5cm, clip]{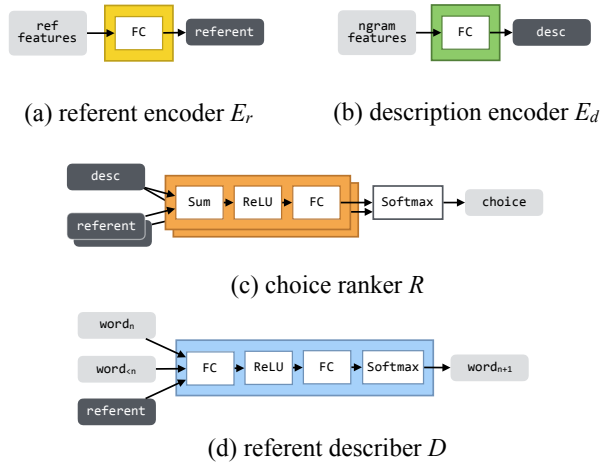}
  \vspace{0.1em}
  \caption{Diagrams of modules used to construct speaker and listener models.
    ``FC'' is a fully-connected layer (a matrix multiply) and ``ReLU'' is a
    rectified linear unit. The encoder modules (a,b) map from feature
    representations (in gray) to embeddings (in black), while the ranker (c) and
    describer modules (d) respectively map from embeddings to 
    decisions
    and strings.}
  \label{fig:modules}
\end{figure}

All listener and speaker models are built from a kit of simple building blocks
for working with multimodal representations of images and text:
\begin{enumerate}
  \item a \textbf{referent encoder} $E_r$
  \item a \textbf{description encoder} $E_d$
  \item a \textbf{choice ranker} $R$
  \item a \textbf{referent describer} $D$
\end{enumerate}
These are depicted in \autoref{fig:modules}, and specified more formally below.
All modules are parameterized by weight matrices, written with capital letters
$W_1$, $W_2$, etc.; we refer to the collection of weights for all modules
together as $W$.

\paragraph{Encoders}
The referent and description encoders produce a linear embedding of referents
and descriptions in a common vector space.
\begin{align}
  &\textrm{Referent encoder:} & E_r(r) = W_1 f(r) \\
  &\textrm{Description encoder:} & E_d(d) = W_2 f(d)
\end{align}

\paragraph{Choice ranker}
The choice ranker takes a string encoding and a collection of referent
encodings, assigns a score to each (string, referent) pair, and then transforms
these scores into a distribution over referents. We write $R(e_i|e_{-i}, e_d)$ for
the probability of choosing $i$ in contrast to the alternative; for example,
$R(e_2|e_1, e_d)$ is the probability of answering ``2'' when presented with
encodings $e_1$ and $e_2$.
\begin{align*}
  s_1 &= w_3^\top \rho(W_4 e_1 + W_5 e_d) \\
  s_2 &= w_3^\top \rho(W_4 e_2 + W_5 e_d) \\
  R(e_i | e_{-i}, e_d) &= \frac{e^{s_i}}{e^{s_1} + e^{s_2}} \numberthis
\end{align*}
(Here $\rho$ is a rectified linear activation function.)

\paragraph{Referent describer}
The referent describer takes an image encoding and outputs a description using a
(feedforward) conditional neural language model.  We express this model as a
distribution $p(d_{n+1} | d_n, d_{<n}, e_r)$, where $d_n$ is an indicator
feature on the last description word generated, $d_{<n}$ is a vector of
indicator features on all other words previously generated, and $e_r$ is a
referent embedding. This is a ``2-plus-skip-gram'' model, with local positional
history features, global position-independent history features, and features on
the referent being described. To implement this probability distribution, we
first use a multilayer perceptron to compute a vector of scores $s$ (one $s_i$
for each vocabulary item): $s = W_6 \rho (W_7 [d_n, d_{<n}, e_i])$. We then
normalize these to obtain probabilities: $p_i = e^{s_i} / \sum_j e^{s_j}$.
Finally,
$
  p(d_{n+1} | d_n, d_{<n}, e_r) = p_{d_{n+1}}
$.

\subsection{Base models}
\label{sec:approach:base}

\begin{figure}
  \centering
  \includegraphics[width=\columnwidth, trim=0cm 11.5cm 11.6cm 0cm, clip]{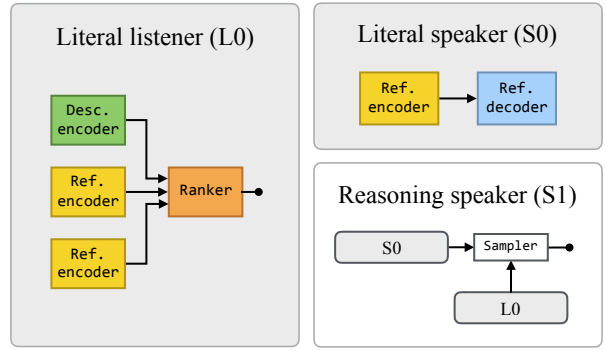}
  \caption{Schematic depictions of models. The literal listener $\lzero$ maps
    from descriptions and reference candidates to reference decisions. The
    literal speaker $\szero$ maps directly from scenes to descriptions, ignoring
    context, while the reasoning speaker uses samples from $\szero$  and scores
  from both $\lzero$ and $\szero$ to produce contextually-appropriate captions.}
  \label{fig:models}
\end{figure}

From these building blocks, we construct a pair of base models. The first of
these is a \textbf{literal listener} $\lzero$, which takes a description and a
set of referents, and chooses the referent most likely to be described. This
serves the same purpose as the base listener in the general derived approach
described in the introduction.  We additionally construct a \textbf{literal
speaker} $\szero$, which takes a referent in isolation and outputs a
description. The literal speaker is used for efficient inference over the space
of possible descriptions, as described in
\autoref{sec:approach:reasoning}.  $\lzero$ is, in essence, a retrieval model,
and $\szero$ is neural captioning model. 

Both of the base models are probabilistic: $\lzero$ produces a distribution over
referent choices, and $\szero$ produces a distribution over strings.
They are depicted with shaded backgrounds in \autoref{fig:models}.

\paragraph{Literal listener}
Given a description $d$ and a pair of candidate referents $r_1$ and $r_2$, the
literal listener embeds both referents and passes them to the ranking module,
producing a distribution over choices $i$.
\begin{align*}
  e_d &= E_d(d) \\
  e_1 &= E_r(r_1) \\
  e_2 &= E_r(r_2) \\
  p_{\lzero}(i|d, r_1, r_2) &= R(e_i | e_{-i}, e_d) \numberthis
\end{align*}
That is, $p_{\lzero}(1|d, r_1, r_2) = R(e_1 | e_2, e_d)$ and vice-versa.
This model is trained contrastively, by solving the following optimization
problem:
\begin{equation}
  \max_W \sum_j \log p_{\lzero}(1 | d_j, r_j, r')
\end{equation}
Here $r'$ is a random distractor chosen uniformly from the training set.  For
each training example $(r_i, d_i)$, this objective attempts to maximize the
probability that the model chooses $r_i$ as the referent of $d_i$ over a random
distractor.

This contrastive objective ensures that our approach is
applicable even when there is not a naturally-occurring source of
target--distractor pairs, as previous work
\cite{Golland10Game,Monroe15RationalSpeech} has required.  Note that this
can also be viewed as a version of the loss described by
\newcite{Smith05Contrastive}, where it approximates a
likelihood objective that encourages $\lzero$ to prefer $r_i$ to every other
possible referent simultaneously.

\paragraph{Literal speaker}
As in the figure, the literal speaker is obtained by composing a referent
encoder with a describer, as follows:
\begin{align*}
  e &= E_r(f(r)) \\
  p_{\szero}(d|r) &= D_d(d|e)
\end{align*}
As with the listener, the literal speaker should be understood as producing a
distribution over strings. It is trained by maximizing the conditional
likelihood of captions in the training data:
\begin{equation}
  \max_W \sum_i \log p_{\szero}(d_i | r_i)
\end{equation}

These base models are intended to be the minimal learned equivalents of the
hand-engineered speakers and hand-written grammars employed in previous
derived approaches \cite{Golland10Game}. The neural encoding/decoding
framework implemented by the modules in the previous subsection provides a
simple way to map from referents to descriptions and descriptions to judgments
without worrying too much about the details of syntax or semantics. Past work
amply demonstrates that neural conditional language models are powerful enough
to generate fluent and accurate (though not necessarily pragmatic) descriptions
of images or structured representations \cite{Donahue15LRCN}. 

\subsection{Reasoning model}
\label{sec:approach:reasoning}

As described in the introduction, the general derived approach to pragmatics
constructs a base listener and then selects a description that makes it behave
correctly. Since the assumption that listeners will behave deterministically is
often a poor one, it is common for such derived approaches to implement
\emph{probabilistic} base listeners, and maximize the probability of correct
behavior.

The neural literal listener $\lzero$ described in the preceding section is such a
probabilistic listener. Given a target $i$ and a pair of candidate referents $r_1$
and $r_2$, it is natural to specify the behavior of a reasoning speaker as
simply:
\begin{equation}
  \label{eq:simple-reasoning}
  \max_d p_{\lzero}(i|d, r_1, r_2)
\end{equation}

At a first glance, the only thing necessary to implement this model is the
representation of the literal listener itself. When the set of possible utterances
comes from a fixed vocabulary \cite{Vogel13Grice} or a grammar small enough to
exhaustively enumerate \cite{Smith13BayesianPragmatics} the operation $\max_d$ in
\autoref{eq:simple-reasoning} is practical. 

For our purposes, however, we would like the model to be capable of producing
arbitrary utterances. Because the score $p_{\lzero}$ is produced by a
discriminative listener
model, and does not factor along the words of the description, there is no
dynamic program that enables efficient inference over the space of all strings.

We instead use a sampling-based optimization procedure. The key ingredient here
is a good \emph{proposal
distribution} from which to sample sentences likely to be assigned high
weight by the model listener. For this we turn to the
literal speaker $\szero$ described in the previous section.
Recall that this speaker produces a distribution over plausible descriptions of
isolated images, while ignoring pragmatic context. We can use it as a source of
candidate descriptions, to be reweighted according to the expected behavior of
$\lzero$.
The full specification of a sampling neural reasoning speaker is as follows:
\begin{enumerate}
  \item Draw samples $d_1, \dots d_n \sim p_{\szero}(\cdot | r_i)$.
  \item Score samples: 
      $p_k = p_{\lzero}(i | d_k, r_1, r_2)$.
  \item Select $d_k$ with $k = \arg\max p_k$.
\end{enumerate}

While primarily to enable efficient inference, we can also use the literal
speaker to serve a different purpose: ``regularizing'' model behavior towards
choices that are adequate and fluent, rather than exploiting strange model
behavior. Past work has restricted the set of utterances in a way that
guarantees fluency. But with an imperfect learned listener model, and a
procedure that optimizes this listener's judgments directly, the speaker model
might accidentally discover the kinds of pathological optima that neural
classification models are known to exhibit \cite{Goodfellow14Pathologies}---in
this case, sentences that cause exactly the right response from $\lzero$, but no
longer bear any resemblance to human language use. To correct this, we allow the
model to consider two questions: as before, ``how likely is it that a listener
would interpret this sentence correctly?'', but additionally ``how likely is it that a
speaker would produce it?''

Formally, we introduce a parameter $\lambda$ that trades off between $\lzero$
and $\szero$, and take the reasoning model score in step 2 above to be:
\begin{equation}
  \label{eq:score}
    p_k = p_{\szero}(d_k | r_i)^\lambda\ \cdot\ 
    p_{\lzero}(i | d_k, r_1, r_2)^{1-\lambda}
  \end{equation}
This can be viewed as a weighted \emph{joint} probability that a sentence is
both uttered by the literal speaker and correctly interpreted by the literal
listener, or alternatively in terms of Grice's conversational maxims
\cite{Grice70Conversation}: $\lzero$ encodes the maxims of \emph{quality} and
\emph{relation}, ensuring that the description contains enough information for
$L$ to make the right choice, while $\szero$ encodes the maxim of \emph{manner},
ensuring that the description conforms with patterns of human language use.
Responsibility for the maxim of \emph{quantity} is shared: $\lzero$ ensures that
the model doesn't say too little, and $\szero$ ensures that the model doesn't
say too much.

\section{Evaluation}
\label{sec:eval}

We evaluate our model on the reference game RG described in the introduction. In
particular, we construct instances of RG using the Abstract Scenes Dataset
introduced by \newcite{Zitnick13Abstract}. Example scenes are shown in
\autoref{fig:teaser} and 
Figure 4.
The dataset contains pictures
constructed by humans and described in natural language.  Scene representations
are available both as rendered images and as feature representations containing
the identity and location of each object; as noted in
\autoref{sec:approach:prelim}, we use this feature set to produce our referent
representation $f(r)$.  This dataset was previously used for a variety of
language and vision tasks (e.g.\ \newcite{Ortiz15Abstract},
\newcite{Zitnick14Abstract}). It consists of 10,020 scenes, each annotated with
up to 6 captions.

The abstract scenes dataset provides a more challenging version of RG than
anything we are aware of in the existing computational pragmatics literature,
which has largely used the {\sc tuna} corpus of isolated object descriptions
\cite{Gatt07TUNA} or
small synthetic datasets \cite{Smith13BayesianPragmatics}. By contrast, the
abstract scenes data was generated by humans looking at complex images with
numerous objects, and features grammatical errors, misspellings, and a
vocabulary an order of magnitude larger than {\sc tuna}. Unlike previous work,
we have no prespecified in-domain grammar, and no direct supervision of the
relationship between scene features and lexemes.



We perform a human evaluation using Amazon Mechanical Turk.
We begin by holding out a development set and a test set; each held-out set
contains 1000 scenes and their accompanying descriptions. For each held-out set,
we construct two sets of 200 paired (target, distractor) scenes: \textbf{All},
with up to four differences between paired scenes, and \textbf{Hard}, with
exactly one difference between paired scenes.  (We take the number of
differences between scenes to be the number of objects that appear in one scene
but not the other.)

We report two evaluation metrics. \emph{Fluency} is determined by showing human
raters isolated sentences, and asking them to rate linguistic quality on a scale
from 1--5. \emph{Accuracy} is success rate at RG: as in \autoref{fig:teaser},
humans are shown two images and a model-generated description, and asked to
select the image matching the description.

In the remainder of this section, we measure the tradeoff between fluency and
accuracy that results from different mixtures of the base models
(\autoref{sec:eval:base}), measure the number of samples needed to obtain good
performance from the reasoning listener (\autoref{sec:eval:samples}), and attempt
to approximate the reasoning listener with a monolithic ``compiled'' listener
(\autoref{sec:eval:compiled}). In \autoref{sec:eval:final} we report final
accuracies for our approach and baselines.

\begin{figure}
  \centering
  \includegraphics[width=\columnwidth, trim=0cm 0cm 0cm 1.5cm, clip]{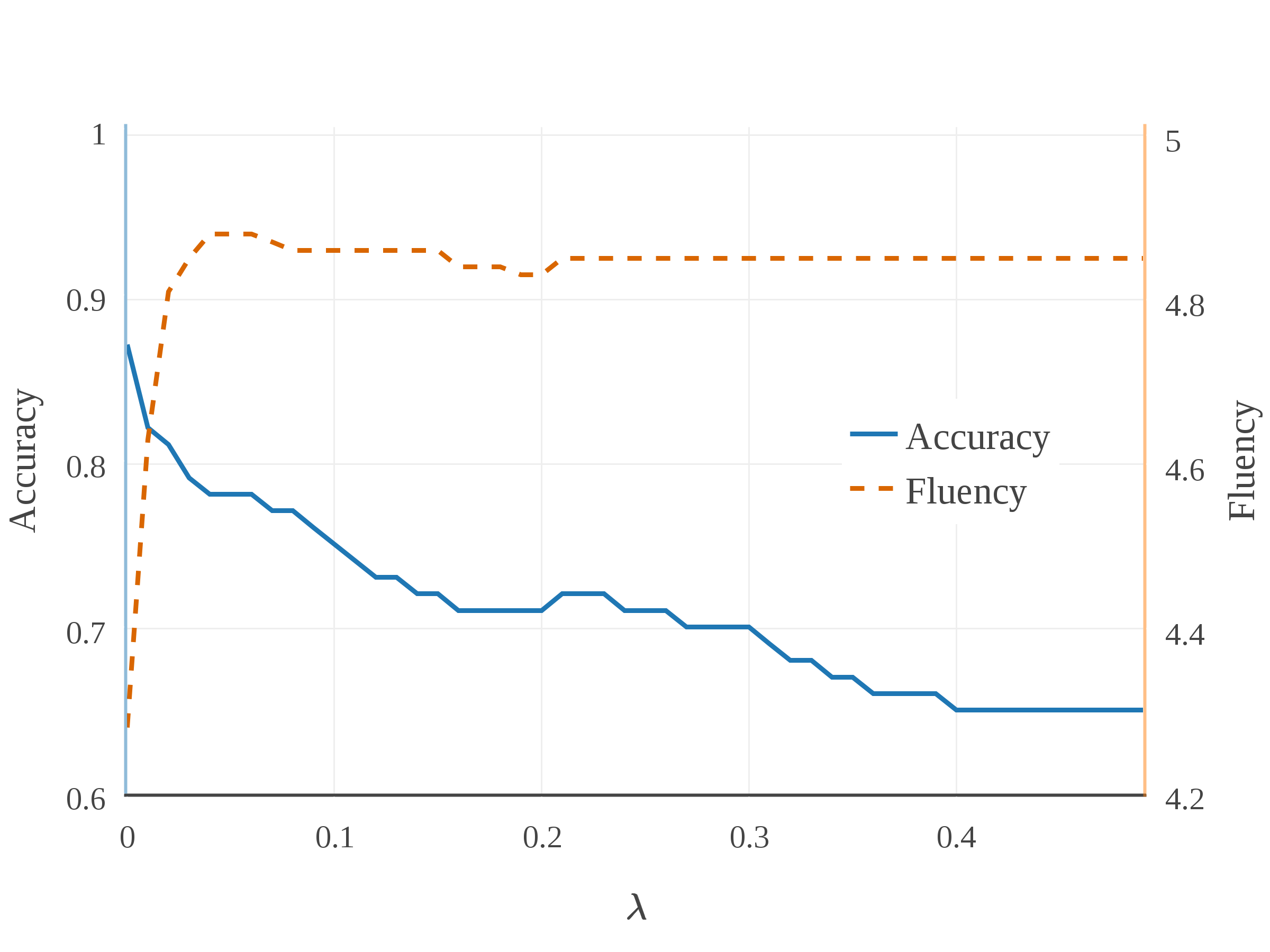}
  \vspace{-.5em}
  \caption*{\footnotesize Figure 5: Tradeoff between speaker and listener models, controlled by the
    parameter $\lambda$ in \autoref{eq:score}. With $\lambda = 0$, all
  weight is placed on the literal listener, and the model produces highly
  discriminative but somewhat disfluent captions. With $\lambda = 1$, all weight
  is placed on the literal speaker, and the model produces fluent but generic
  captions.}
  \label{fig:tradeoff}
\end{figure}

\begin{table}[b]
  \centering
  \footnotesize
  \begin{tabular}{c|cccc}
    \toprule
    \# samples & 1   & 10  & 100 & 1000 \\
    Accuracy (\%)  & 66  & 75  & 83  & 85 \\
    \bottomrule
  \end{tabular}
  \vspace{0.5em}
  \caption{$\sone$ accuracy vs.\ number of samples.}
  \label{tab:nsamples}
\end{table}

\subsection{How good are the base models?}
\label{sec:eval:base}

To measure the performance of the base models, we draw 10 samples $d_{jk}$ for a
subset of 100 pairs $(r_{1,j}, r_{2,j})$ in the Dev-All set. We collect human
fluency and accuracy judgments for each of the 1000 total samples. This allows
us to conduct a post-hoc search over 
values of
$\lambda$: for a range of $\lambda$, we compute the average accuracy and fluency
of the highest scoring sample. By varying $\lambda$, we can view the tradeoff
between accuracy and fluency that results from interpolating between the
listener and speaker model---setting $\lambda = 0$ gives samples from
$p_{\lzero}$, and $\lambda = 1$ gives samples from $p_{\szero}$.

Figure 5 shows the resulting accuracy and fluency for various values of
$\lambda$. It can be seen that relying entirely on the
listener gives the highest accuracy but degraded fluency. However,
by adding only a very small weight to the speaker model, it is possible to
achieve near-perfect fluency without a substantial decrease in accuracy.
Example sentences for an individual reference game are shown in
\autoref{fig:gradient}; increasing
$\lambda$ causes captions to become more generic.
For the
remaining experiments in this paper, we take $\lambda = 0.02$, finding that this
gives excellent performance on both metrics.

On the development set, $\lambda=0.02$ results in an \textbf{average fluency of 4.8}
(compared to 4.8 for the literal speaker $\lambda=1$). This high fluency can be
confirmed by inspection of model samples 
(Figure 4). We thus focus
on \textbf{accuracy} or the remainder of the evaluation.

\begin{figure*}
\footnotesize
\center
\begin{tabular}{ccc}
  \includegraphics[cfbox=highlight 4mm 0mm,width=0.37\columnwidth]{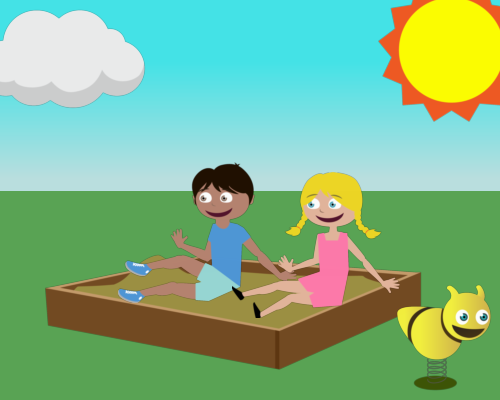} 
  \includegraphics[cfbox=white 4mm 0mm,width=0.37\columnwidth]{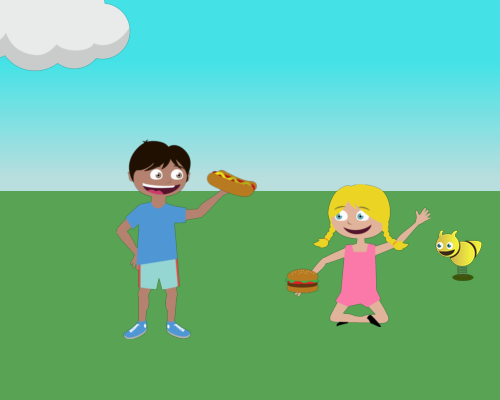} &\strut\hspace{3em}\strut&
  \includegraphics[cfbox=highlight 4mm 0mm,width=0.37\columnwidth]{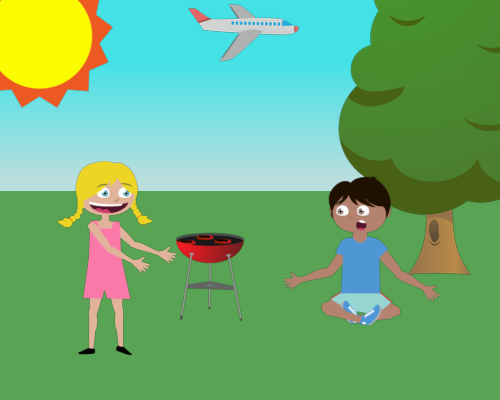} 
  \includegraphics[cfbox=white 4mm 0mm,width=0.37\columnwidth]{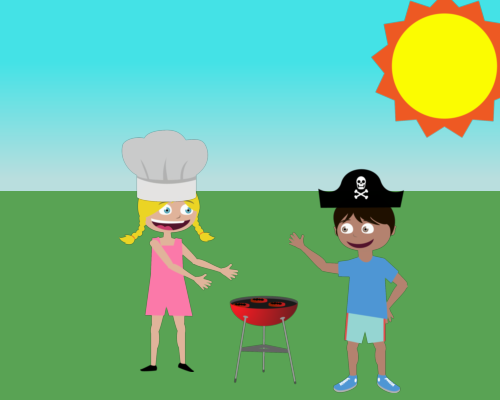} \\[0.5em]
  (a) \it the sun is in the sky &&
  (d) \it the plane is flying in the sky \\

  [contrastive] &&
  [contrastive] \\[2em]
  \includegraphics[cfbox=highlight 4mm 0mm,width=0.37\columnwidth]{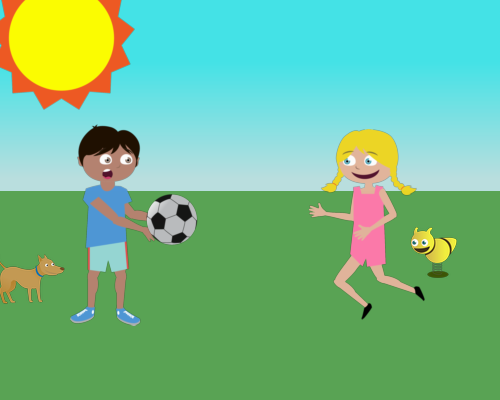}
  \includegraphics[cfbox=white 4mm 0mm,width=0.37\columnwidth]{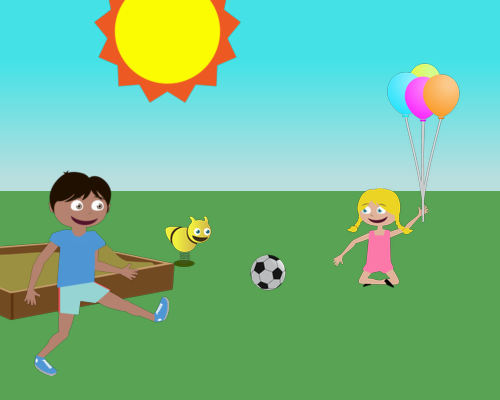} &&
  \includegraphics[cfbox=highlight 4mm 0mm,width=0.37\columnwidth]{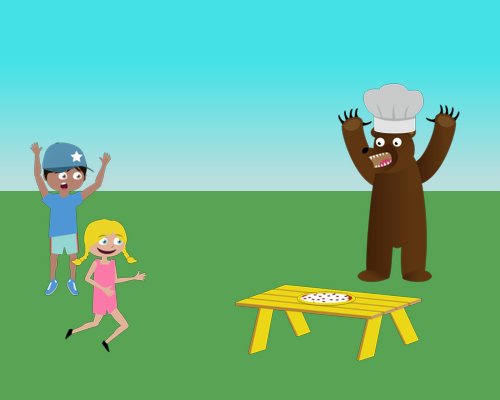}
  \includegraphics[cfbox=white 4mm 0mm,width=0.37\columnwidth]{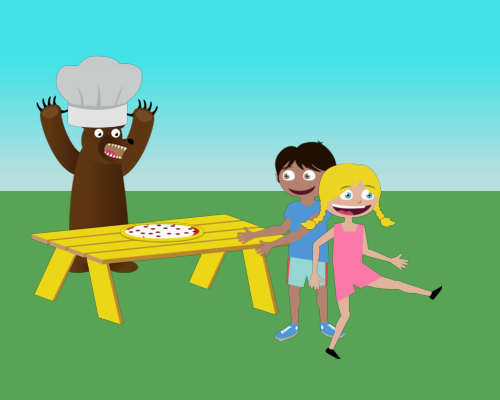}
  \\[0.5em]
  (c) \it the dog is standing beside jenny &&
  (b) \it mike is wearing a chef's hat \\

  [contrastive] &&
  [non-contrastive] \\
\end{tabular}
  \vspace{0.5em}
\caption{Figure 4:
  Four randomly-chosen samples from our model. For each, the target
image is shown on the left, the distractor image is shown on the right, and
description generated by the model is shown below. All descriptions are
fluent, and generally succeed in uniquely identifying the target scene, even
when they do not perfectly describe it (e.g. (c)). 
These samples are broadly
representative of the model's performance (\autoref{tab:results}).
}
\label{fig:samples}
\end{figure*}

\subsection{How many samples are needed?}
\label{sec:eval:samples}

Next we turn to the computational efficiency of the reasoning model. As in all
sampling-based inference, the number of samples that must be drawn from the
proposal is of critical interest---if too many samples are needed, the model
will be too slow to use in practice. Having fixed $\lambda=0.02$ in the
preceding section, we measure accuracy for versions of the reasoning model that
draw 1, 10, 100, and 1000 samples. Results are shown in \autoref{tab:nsamples}. We find
that gains continue up to 100 samples.

\begin{figure}[b!]
  \centering
  \small
  \hfill
  \includegraphics[cfbox=highlight 4mm 0mm,width=0.4\columnwidth]{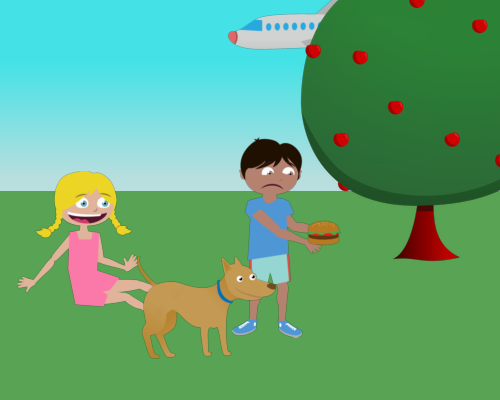}
  \hfill
  \includegraphics[cfbox=white 4mm 0mm,width=0.4\columnwidth]{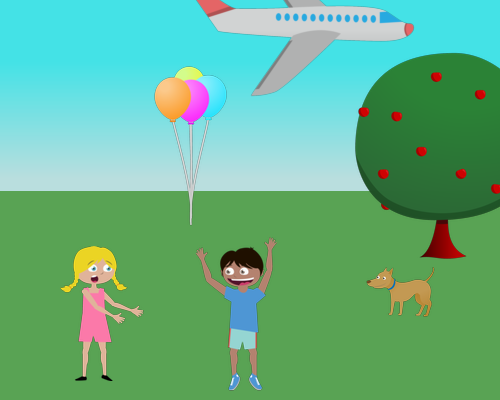}
  \hfill
  \strut
  \\
  {\footnotesize \strut \hspace{1.4cm} (a) target \hspace{2cm} (b) distractor
  \hfill \strut }\\[1em]

  \begin{samplebox} 
    \footnotesize
    \begin{tabular}{lll}
      (prefer $\lzero$) & \tt 0.0 & \emph{a hamburger on the ground} \\[0.2em]
                    & \tt 0.1 & \emph{mike is holding the burger} \\[0.2em]
      (prefer $\szero$) & \tt 0.2 & \emph{the airplane is in the sky}
    \end{tabular}\\[-0.2em]
  \end{samplebox}\vspace{0.5em}

  \caption{Captions for the same pair with varying $\lambda$. Changing $\lambda$
  alters both the naturalness and specificity of the output.}
  \label{fig:gradient}
\end{figure}

\subsection{Is reasoning necessary?}
\label{sec:eval:compiled}

Because they do not require complicated inference procedures, direct approaches
to pragmatics typically enjoy better computational efficiency than derived
ones. Having built an accurate derived speaker, can we bootstrap
a more efficient direct speaker?

To explore this, we constructed a ``compiled'' speaker model as follows: Given
reference candidates $r_1$ and $r_2$ and target $t$, this model produces
embeddings $e_1$ and $e_2$, concatenates them together into a ``contrast
embedding'' $[e_t, e_{-t}]$, and then feeds this whole embedding into a string
decoder module. Like $\szero$, this model generates captions without the need
for discriminative rescoring; unlike $\szero$, the contrast embedding means
this model can in principle learn to produce pragmatic captions, if given
access to pragmatic training data. Since no such training data exists, we train
the compiled model on captions sampled from the reasoning speaker itself.

This model is evaluated in \autoref{tab:compiled}. While the distribution of
scores is quite different from that of the base model (it improves noticeably
over $\szero$ on scenes with 2--3 differences), the overall gain is negligible
(the difference in mean scores is not significant). The compiled model
significantly underperforms the reasoning model. These results suggest either
that the reasoning procedure is not easily approximated by a shallow neural
network, or that example descriptions of randomly-sampled training pairs (which
are usually easy to discriminate) do not provide a strong enough signal for a
reflex learner to recover pragmatic behavior.

\newcommand{\whdashline}{\\[-0.8em]\hdashline\\[-0.8em]}

\begin{table}
  \centering
  \footnotesize
  \begin{tabular}{lcccc}
    \toprule
    & \multicolumn{2}{c}{Dev acc.\ (\%)} & \multicolumn{2}{c}{Test acc.\
    (\%)} \\
    \cmidrule(lr){2-3} \cmidrule(lr){4-5}
    Model & All & Hard & All & Hard \\
    \midrule
    Literal ($\szero$)  & 66     & 54     & 64     & 53 \\
    Contrastive         & 71     & 54     & 69     & 58 \\
    Reasoning ($\sone$) & \bf 83 & \bf 73 & \bf 81 & \bf 68 \\
    \bottomrule
  \end{tabular}
  \vspace{0.5em}
  \caption{Success rates at RG on abstract scenes.
    ``Literal'' is a captioning
    baseline corresponding to the base speaker $\szero$.
    ``Contrastive'' is a reimplementation of the approach of
    Mao et al. (2015).
    ``Reasoning'' is the model from this paper.
    All differences between our model and baselines are significant
    ($p < 0.05$, Binomial).}
  \label{tab:results}
\end{table}

\begin{table}
  \centering
  \footnotesize
  \begin{tabular}{lccccc}
    \toprule
    & \multicolumn{4}{c}{\# of differences} \\
    & 1 & 2 & 3 & 4 & Mean \phantom{(\%)} \\
    \midrule
      Literal ($\szero$)  & 50 & 66 & 70 & 78 & 66 (\%) \\
      Reasoning           & 64 & 86 & 88 & 94 & 83 \phantom{(\%)} \\
      Compiled ($\sone$)  & 44 & 72 & 80 & 80 & 69 \phantom{(\%)} \\
    \bottomrule
  \end{tabular}
  \vspace{0.5em}
  \caption{Comparison of the ``compiled'' pragmatic speaker model with literal
  and explicitly reasoning speakers. The models are evaluated on subsets of the
  development set, arranged by difficulty: column headings indicate the number of
  differences between the target and distractor scenes. 
  }
  \label{tab:compiled}
\end{table}

\subsection{Final evaluation}
\label{sec:eval:final}

Based on the following sections, we keep $\lambda = 0.02$ and use 100 samples to
generate predictions. We evaluate on the test set, comparing this
\textbf{Reasoning} model $\sone$ to two baselines: \textbf{Literal}, an image
captioning model trained normally on the abstract scene captions (corresponding
to our $\lzero$), and \textbf{Contrastive}, a model trained with a soft
contrastive objective, and previously used for visual referring expression
generation \cite{Mao15Generation}.

\begin{figure}
  \centering
  \footnotesize
  \includegraphics[cfbox=white 4mm 0mm,width=0.33\columnwidth]{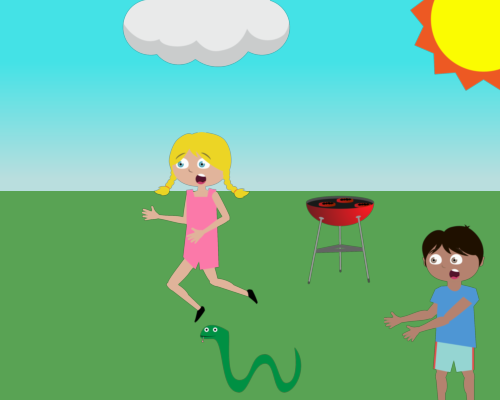}%
  \includegraphics[cfbox=highlight 4mm 0mm,width=0.33\columnwidth]{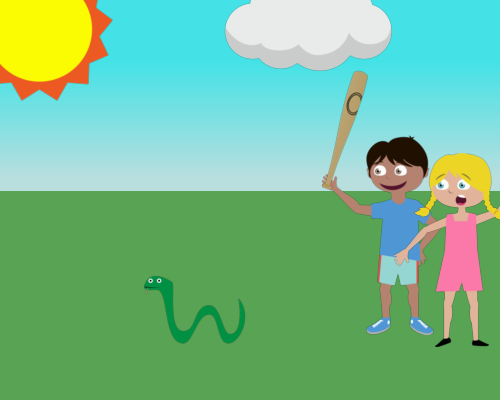}%
  \includegraphics[cfbox=white 4mm 0mm,width=0.33\columnwidth]{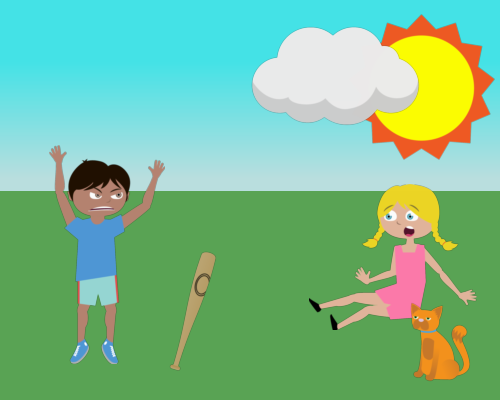} \\
  \hfill (a) \hfill\hfill (b) \hfill\hfill (c) \hfill \strut \\[0.5em]
  \hspace{-.5em}
  \begin{tabular}{ll}
  (b vs.\ a) & \emph{mike is holding a baseball bat} \\
  (b vs.\ c) & \emph{the snake is slithering away from mike and jenny}
  \end{tabular}
  \vspace{0.5em}
  \caption{Descriptions of the same image in different contexts. When the target
    scene (b) is contrasted with the left (a), the system describes a bat; when the
    target scene is contrasted with the right (c), the system describes a
    snake.}
    \label{fig:context}
\end{figure}

Results are shown in \autoref{tab:results}. Our reasoning model outperforms both
the literal baseline and previous work by a substantial margin, achieving an
improvement of 17\% on all pairs set and 15\% on hard pairs.\footnote{
  For comparison, a model with hand-engineered pragmatic behavior---trained
  using a feature representation with indicators on only those objects that
  appear in the target image but not the distractor---produces an accuracy of
  78\% and 69\% on all and hard development pairs respectively. In addition to
  performing slightly worse than our reasoning model, this alternative
  approach relies on the structure of scene representations and cannot be
  applied to more general pragmatics tasks.
}
Figures 4
and \ref{fig:context} show various representative
descriptions from the model.

\section{Conclusion}

We have presented an approach for learning to generate pragmatic descriptions
about general referents, even without training data collected in a pragmatic
context. Our approach is built from a pair of simple neural base models, a
listener and a speaker, and a high-level model that reasons about their outputs
in order to produce pragmatic descriptions. In an evaluation on a standard
referring expression game, our model's descriptions produced correct behavior in
human listeners significantly more often than existing baselines.

It is generally true of existing derived approaches to pragmatics that much of
the system's behavior requires hand-engineering, and generally true of direct
approaches (and neural networks in particular) that training is only possible
when supervision is available for the precise target task. By synthesizing these
two approaches, we address both problems, obtaining pragmatic behavior without
domain knowledge and without targeted training data. We believe that this
general strategy of using reasoning 
to obtain novel contextual
behavior from neural decoding models might be more broadly applied.

%

\bibliographystyle{acl2016}
\bibliography{jacob}

\end{document}